\documentclass{article}
\usepackage{ijcai24}

\usepackage{times}
\usepackage{soul}
\usepackage{url}
\usepackage[hidelinks]{hyperref}
\usepackage[utf8]{inputenc}
\usepackage[small]{caption}
\usepackage{graphicx}
\usepackage{amsmath}
\usepackage{amsthm}
\usepackage{booktabs}
\usepackage{algorithm}
\usepackage{algorithmic}
\usepackage[switch]{lineno}

\usepackage{tabularx}
\usepackage{makecell}
\usepackage{multirow}
\usepackage{bigstrut}
\usepackage{threeparttable}
\usepackage{color}
\usepackage{marvosym}

\title{Guiding Clinical Reasoning with Large Language Models via Knowledge Seeds}

\author{
    Jiageng Wu\textsuperscript{1}, Xian Wu\textsuperscript{2}, Jie Yang\textsuperscript{3}\\
    \affiliations
    \textsuperscript{1}Zhejiang University, Hangzhou, China \\
    \textsuperscript{2}Tencent Youtu Lab, Jarvis Research Center, Beijing, China\\
    \textsuperscript{3}Brigham and Women's Hospital, Harvard Medical School, Boston, USA \\
    \emails
    jiagengwu@zju.edu.cn, kevinxwu@tencent.com, jieynlp@gmail.com
}

\begin{document}

\maketitle

\begin{abstract}

% Clinical reasoning refers to the cognitive process that physicians conduct in evaluating and managing patients. Typical clinical reasoning tasks consist of suggesting necessary examinations, diagnosing patients' diseases and deciding proper therapies etc. Accurate clinical reasoning requires extensive medical knowledge and rich clinical experience, setting high bar for physicians. However, accessing high quality medical care is difficult due to the excessive patient number and limited physician resources. This is even worse in developing countries. Therefore an automatic approach that can assist in clinical reasoning is urged. Recently, the emergence of large language models like ChatGPT and GPT-4 demonstrates their potentials in clinical reasoning. However, these LLMs suffers from the hallucination problems and the generated reasoning content may not align with clinical evidence. In this study, we introduce a novel framework called In-Context Padding (ICP) to enhance LLMs with medical knoweledge. In particular, we infer the critical knowledge keypoint (refering to knowledge seeds) and use these anchors to guide the reasoning process of LLMs. Experiments conducted on two clinical question datasets indicate that ICP markedly improves the reasoning ability of LLMs, particularly when the LLMs encode insufficient knowledge. We also validate the proposed  ICP framework offers a novel approach to enhance LLM performance in non-English and healthcare domains, consequently bridging the capability disparity of LLMs in low-resource applications. 

Clinical reasoning refers to the cognitive process that physicians employ in evaluating and managing patients. This process typically involves suggesting necessary examinations, diagnosing patients' diseases, and selecting appropriate therapies, etc. 
Accurate clinical reasoning requires extensive medical knowledge and rich clinical experience, setting a high bar for physicians. This is particularly challenging in developing countries due to the overwhelming number of patients and limited physician resources, contributing significantly to \textit{global health inequity} and necessitating automated clinical reasoning approaches. Recently, the emergence of large language models (LLMs) such as ChatGPT and GPT-4 have demonstrated their potential in clinical reasoning. However, these LLMs are prone to hallucination problems, and the reasoning process of LLMs may not align with the clinical decision pathways of physicians. In this study, we introduce a novel framework, In-Context Padding (ICP), to enhance LLMs reasoning with medical knowledge. Specifically, we infer critical clinical reasoning elements (referred to as knowledge seeds) and use these as anchors to guide the generation process of LLMs. 
Experiments on two clinical question datasets validate that ICP significantly improves the clinical reasoning ability of LLMs.

\end{abstract}

\section{Introduction}
Clinical reasoning is a pivotal process where healthcare professionals incorporate clinical evidence and medical knowledge to assess, diagnose, and decide on treatment for patients \cite{medical-reasoning-book}. It entails a series of cognitive tasks, including gathering patient information, formulating and evaluating diagnostic hypotheses, and making treatment decisions \cite{clinical-reasoning-definition}. 
Consequently, clinical reasoning requires extensive medical knowledge and rich clinical experience, setting high expectations for physicians. However, in low- and middle-income countries (LMICs), medical resources are often scarce, making it difficult to access high-quality medical care.
Despite accounting for 90\% of the global burden of disease, LMICs only contribute to 12\% of global health spending \cite{lmic-burden}. 
% High-income countries spend about 100 times more on health per capita than low-income countries (\$3039 versus \$30) \cite{lmic-burden}. 
Furthermore, in 98 countries across Asia and Africa, the population of physicians does not meet the minimum threshold required for achieving 80\% universal health coverage, underscoring a critical shortage of qualified clinical specialists \cite{physician-africa-asia-lancet}.

The shortage of healthcare resources urges the emergence of automated approaches with reliable clinical reasoning capabilities to support clinical decisions.
Recently, Large Language Models (LLMs) 
% exhibit capabilities in comprehending humans' intentions and generating coherent responses \cite{llm-survey-2023}. Additionally, their text-to-text architecture enables direct interaction via textual prompts \cite{llm-gpt-unilm}, making it accessible and user-friendly for both physicians and patients. Consequently, they 
have shown great potential in the medical domain \cite{chagpt-review-NEJM-2023}, such as medical education \cite{chatgpt-education-2023}, online consultation \cite{online-medkp}, and clinical report summarization \cite{chatgpt-summary-2023}. Encouragingly, several advanced LLMs qualified for the medical licensing examinations at high scores, such as Med-PaLM \cite{exam-med_palm-2023}, ChatGPT \cite{exam-chatgpt-2023}, and GPT-4 \cite{exam-gpt4-2023}, indicating remarkable proficiency in medical knowledge and clinical case analysis. 

Despite the notable capabilities in comprehending humans' intentions and generating coherent responses \cite{llm-survey-2023}, directly applying LLMs to the medical field has also raised concerns over the generations of incorrect knowledge and hallucination during clinical reasoning \cite{chatgpt-suggestion-eye,chatgpt-language-disparity}. This primarily stems from these advanced LLMs being predominantly trained on general-domain data \cite{review-clinical-dataset}. Lacking extensive training in domain-specific text, they fail to encode sufficient medical expertise and comprehend medical texts laden with specialized concepts \cite{encode-2022}. 
Absent a solid foundation of medical knowledge, LLMs struggle to grasp the intricate medical context and identify critical concerns behind it, making it difficult to generate comprehensive medical inferences.
% fully \cite{llm-adoption-medical}
% Due to the current advanced LLM being English-centric \cite{llm-english-centric}, this insufficiency in data and training is exacerbated in non-English medical scenarios, resulting in performance decline \cite{xiaocong-2023} and biased result \cite{chatgpt-resource-lmic}. 

To tackle these challenges, we propose a novel framework \textbf{I}n-\textbf{C}ontext \textbf{P}adding (\textbf{ICP}) to enhance the inference capacity of LLM in the context of clinical reasoning. ICP consists of four major steps: 1) ICP firstly extracts medical entities from the clinical context and the reasoning objective; 2) In cooperation with the knowledge graph (KG), ICP then infer relevant medical entities (referred to as knowledge seeds) which could be helpful in clinical reasoning; 3) The acquired knowledge seeds are padded to the prompt and used to guide the inference process of LLMs; 4) finally, LLMs generate the clinical reasoning results as well the detailed explanation of how this reasoning is conducted.
% discerns potentially pivotal considerations given the clinical question and enriches the prompt with these knowledge seeds, which direct LLMs to engage in pertinent discussions and gradually refine inference. 
Extensive experiments and analyses on two datasets highlight a significant improvement in both the accuracy and interpretability of LLM. 
The ICP framework incorporates KG and in-context learning of LLM to efficiently bridge the knowledge gap in medical scenarios, ensuring its broad applicability in specialized domains. In addition to clinical reasoning results, the proposed ICP also provides an explanation of the reasoning process. Overall, the contributions are as follows:

\begin{itemize}
    \item We propose \textbf{I}n-\textbf{C}ontext \textbf{P}adding (ICP) which enhances LLMs to conduct clinical reasoning. This is especially beneficial for less developed countries where high-quality medical care is hard to access.
    \item We infer the knowledge seeds from context information which are used as anchors for LLMs to conduct reasoning. This helps to align the LLM generation with the clinical reasoning process of physicians.
    \item Experimental results on two datasets validate the effectiveness of the proposed ICP. In addition to the final answer, ICP also provides a description of the reasoning process, making it more transparent and understandable.
\end{itemize}

% Nonetheless, they primarily focused on improving accuracy and neglected the in-depth investigation for medical reasoning. Convincing evidence and detailed inference for explainable predictions are essential for clinical practice \cite{chatgpt-med-NatMed}, especially in settings with limited medical resources \cite{chatgpt-resource-lancet-2023}. Clinical predictions without clear explanations can be challenging for inexperienced physicians to discern, potentially leading to severe outcomes for patients.

% Moreover, directly applying LLMs in the medical field has also raised concerns over the generations of incorrect knowledge and hallucination during clinical reasoning \cite{chatgpt-suggestion-eye}. This primarily stems from these advanced LLMs being predominantly trained on general-domain data. Lacking extensive training in domain-specific text, they fail to encode sufficient medical expertise and comprehend medical texts laden with specialized concepts \cite{encode-2022}. Due to the current advanced LLM being English-centric \cite{llm-english-centric}, this insufficiency in data and training is exacerbated in non-English medical scenarios, resulting in performance decline \cite{xiaocong-2023} and biased result \cite{chatgpt-resource-lmic}. 
% Absent a solid foundation of medical knowledge, LLMs struggle to fully grasp the intricate medical context and identify critical concerns behind it,  making it difficult to generate comprehensive medical inferences.
% significant, \cite{data-imbalance-2023}

\begin{figure*}[ht]
    \centering
    \includegraphics[width=\linewidth]{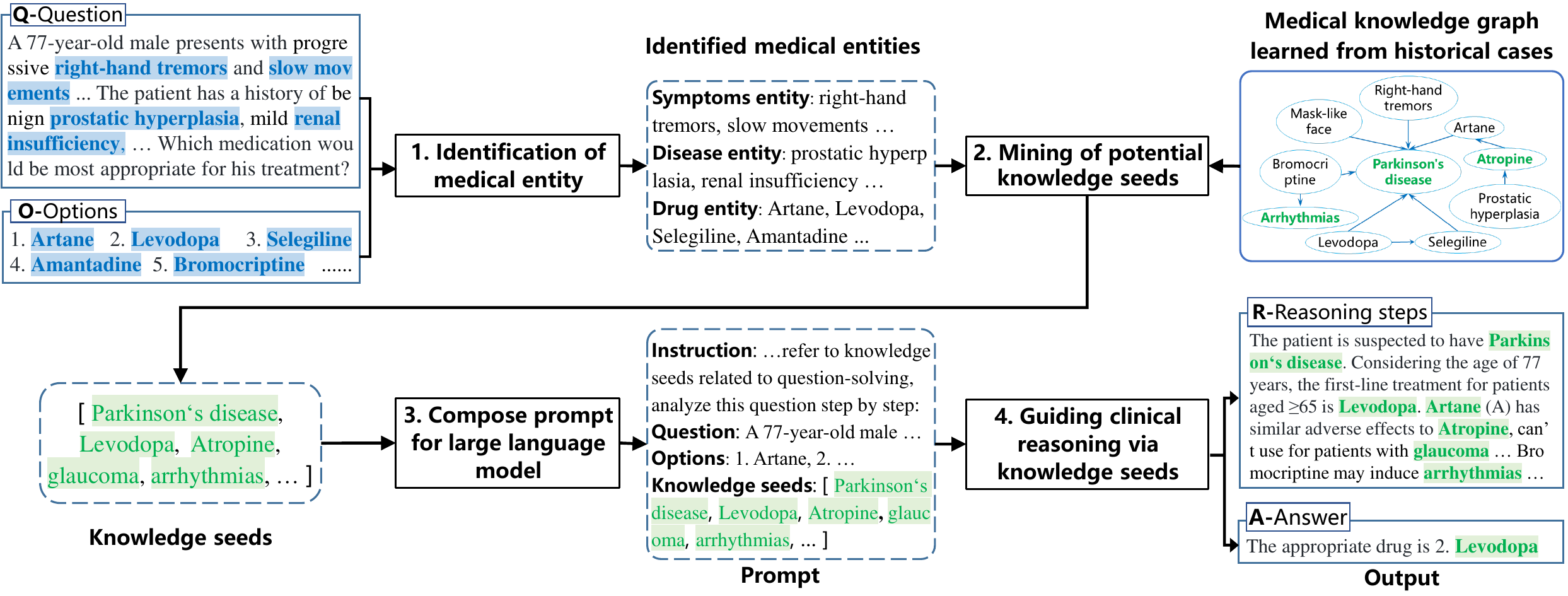}
    \vspace{-0.5cm} 
    \caption{Overall Workflow of In-context Padding. It includes four main steps: 
    1) Identification of medical entities from a question and its candidate options; 
    2) Mine potential knowledge seeds by incorporating the identified medical entities and a medical knowledge graph learned from historical cases;
    3) Compose the whole prompt for LLM, including task instruction, question, options, and mined knowledge seeds;
    4) Guide LLMs to conduct clinical reasoning, leveraging knowledge seeds as anchors to perform relevant inferences and arrive at a conclusion.
    The blue text highlights key medical concepts in the question and options, while the green text indicates the identified knowledge seeds not in the question but critical for the reasoning process. 
    }
    \label{fig: Overall}
    \vspace{-0.2cm} 
\end{figure*}

% To tackle these challenges, we propose a novel framework \textbf{I}n-\textbf{C}ontext \textbf{P}adding (ICP) to enhance the inference capacity of LLM. 
% In cooperation with the knowledge graph (KG), ICP discerns potentially pivotal considerations given the clinical question and enriches the prompt with these knowledge seeds, which direct LLMs to engage in pertinent discussions and gradually refine inference. Extensive experiments and analyses on Chinese medical licensing examinations highlight a significant improvement in both the accuracy and interpretability of LLM. 
% The ICP framework incorporates KG and in-context learning of LLM to efficiently bridge the knowledge gap in non-English medical scenarios, ensuring its broad applicability across various languages and specialized domains. 
% \begin{itemize}
%     \item We construct a medical knowledge graph based on historical clinical questions, which depicts the interrelationship among medical entities and their structure information
% \end{itemize}e

\section{Related Work}
% \subsection{Large language model}
% In recent years, amidst the rapid augmentation in the volume of training data, the scale of model parameters, and the availability of computational resources \cite{scale-llm-2020}, there has been substantial advancement in language models. Since the emergence of BERT in 2018 (0.34B parameters) \cite{transformer-2017}, there has been a consistent trend towards larger models. This is evident in models like GPT-2 \cite{llm-gpt2-2019}, with 1.5B parameters, and GPT-3 \cite{llm-gpt3-2020}, with an amazing 175B parameters. With the rising popularity of ChatGPT, the LLMs have progressively served as the foundation model in AI research \cite{foudation-review-2021}.

% Through extensive pre-training, these LLMs exhibit remarkable ability in text generation and possess rich knowledge \cite{llm-survey-2023}. By utilizing large-scale Supervised Fine-tuning (SFT) across various tasks, they demonstrate their adeptness in intention recognition and solving multi-tasks \cite{instruction-tuning-2022}. Furthermore, Reinforcement Learning from Human Feedback (RLHF) enables these models to provide anthropomorphic responses that align with human values \cite{RLHF-training-2017}.
% Consequently, they have been widely applied across various domains \cite{llm-practice-2023}. 

\subsection{LLMs in Medicine}
% mostly evaluation, some worked on incorporating external knowledge, required high quality structured knowledge e.g. knowledge graph, and has limited application field.  No research focuses on improving inference
As several advanced LLMs have passed the medical licensing examination, there is a growing interest among researchers to explore the deployment of LLMs in clinical environments.
\citeauthor{chatgpt-suggestion-eye}(\citeyear{chatgpt-suggestion-eye}) and \citeauthor{chatgpt-consulation-2023}(\citeyear{chatgpt-consulation-2023}) compared the responses of ChatGPT with medical experts to healthcare questions posed by patients, which indicated the LLM could potentially offer helpful suggestions across various patient inquiries. Moreover, the LLM's responses were also rated significantly higher in terms of both quality and empathy. 
% those of ophthalmologists to eye care questions posed on an online advice forum, which indicated the LLM could potentially offer ophthalmic advice of comparable quality across a spectrum of patient inquiries. A separate study by Ayers et al.\cite{chatgtpt-response-empathy} evaluated the responses of physicians and ChatGPT to questions patients publicly posed on a social media forum. They observed that the ChatGPT's responses were not only preferred over those of the physicians but were also rated significantly higher in terms of both quality and empathy. 
For multi-modality tasks in medicine, ChatCAD \cite{clinical-chatcad-2023} was developed to incorporate the LLMs for an interactive computer-aided diagnosis of medical images. \citeauthor{clinical-radiology-swallow}(\citeyear{clinical-radiology-swallow}) and \citeauthor{clinical-radiology-translating}(\citeyear{clinical-radiology-translating}) assessed the capabilities of LLMs in translating radiology reports into plain language to be easier understood by patients. 

However, the LLM may provide false knowledge and hallucinations due to the misleading brought by the training corpus \cite{chatgpt-suggestion-eye,llm-survey-2023}. This makes them challenging to deploy in a clinical setting, which necessitates rich medical knowledge and experience to conduct rigorous reasoning. While several studies attempted to integrate knowledge to enhance performance in medical tasks \cite{exam-chatgpt-2023,med-kg-2023}, they typically required high-quality, large-scale, and structured medical knowledge for retrieval, which in turn limits their application. 
Furthermore, many of them focus on accuracy, overlooking the desire to improve medical inference during problem-solving, a crucial aspect of integrating LLMs into clinical practice.

\subsection{Reasoning with LLMs}
% chain of thought, Multi-step reasoning, considering add Tree of thought and Graph of thought
Owing to the exceptional ability in text generation and in-context learning, \citeauthor{cot-first-wei2022} (\citeyear{cot-first-wei2022}) elicited the LLM to generate a detailed reasoning process before offering answers to questions under a few-shot setting, called Chain-of-Thought (CoT). It enhanced both accuracy and interpretability in various reasoning tasks. Then, \citeauthor{cot-self-consistency-wang2022}(\citeyear{cot-self-consistency-wang2022})explored the potential of sampling multiple reasoning paths, combining them using an ensemble voting technique to bolster performance. Meanwhile, \citeauthor{cot-step-kojima2022}(\citeyear{cot-step-kojima2022}) introduced a simple prompt, "\textit{Let's think step by step}", to prompt LLMs to elucidate their analysis and then conclude with the answer, without any manually crafted examples.

Addressing the challenges of complex multi-step reasoning, numerous novel frameworks were developed to enhance the logical reasoning ability of the LLMs. \citeauthor{cot-react-yao2022}(\citeyear{cot-react-yao2022}) decomposed the entirety of task-solving into reasoning and acting steps to progressively complete it. Taking a different approach, the Tree of Thought (ToT) \cite{cot-tot-2023} empowered LLMs to consider multiple different reasoning paths and self-evaluate choices to decide the next course of action, achieving notable performance improvements in tasks like the Game of 24 and word-based games. Furthermore, the Graph of Thought (GoT) \cite{cot-got-2023} formulates the reasoning process using a graph to enhance the problem-modeling capability of CoT. 
% emulate the human reasoning process, thus enhancing 
Nonetheless, these studies mainly concentrated on mathematical reasoning and word games. There is limited exploration in the medical domain, where problem-solving also requires rigorous reasoning.
% Similarly, the Least-to-Most \cite{cot-least-to-most-zhou2022} segmented a complex problem into several simpler subproblems. These subproblems were then solved sequentially, where the solution to each subproblem was facilitated by the answers to the preceding ones. 

\section{Methodology}

\subsection{Problem Formulation}
In this subsection, we formulate the clinical reasoning task. We denote an instance of clinical reasoning with four elements: $\{Q, O, R, A\}$, where $Q$ refers to the context information and the reasoning objective, $O=\{o_0, ..., o_s\}$ refers to $s$ candidate options, $A$ refers to the correct answer which is one option in $O$, and $R$ refers to the detailed analysis to conclude $A$. 
Given $Q$ and options $O$, conducting clinical reasoning can be formulated as estimating the probability of generating reasoning steps and then determining the correct answer $P(R, A|Q, O)$. 
As shown in Figure \ref{fig: Overall}, $Q$ refers to 1) the conditions of this patient; 2) the reasoning objective which is to infer the most appropriate treatment; $O$ refers to five candidate drugs for treatments: Artane, Levodopa, Selegiline, Amantadine and Bromocriptine; $A$ refers to the best choice Levodopa; $R$ refers to the description of reasoning process. 

\subsection{Overall Workflow}
% To enhance the model's ability in question-solving and analysis generation, we guide the LLM to generate a detailed analysis with the guidance of key medical entities, which can be determined by the context of the question and a medical KG. 
Figure \ref{fig: Overall} illustrates the overall workflow of ICP.
% \footnote{All prompts and questions in this study were written in Chinese to be consistent with actual CNMLE, we translated them into English in the paper for better readability.}. 
The fundamental principle of ICP is to identify the most relevant knowledge seeds and incorporate them into the in-context prompt as potential anchors. These anchors guide the construction of a multi-step inference path. The process comprises four main steps: 
1) Identification of medical entities to understand the context information and the primary object for reasoning; 
2) Mining of the potential knowledge seeds by incorporating identified medical entities and a medical knowledge graph learned from historical cases;
3) Composition of the whole prompt for LLM, including task instruction, context, options, and mined knowledge seeds;
4) Guiding clinical reasoning with LLM via knowledge seeds, which serve as anchors to conduct relevant inferences and conclude the final answer.

\subsection{Identification of Medical Entities}
For each instance, we concatenate $Q$ and $O$ and then extract the medical entities, including disease, symptoms, drugs, and any medical concepts, which briefly but precisely represent the medical context and the reasoning target. For the training instances, we also extract the medical entities discussed in its detailed analysis $R$. 

To effectively extract the medical entities, we harness the exceptional in-context learning capabilities of LLMs and few-shot learning. Specifically, we provide the LLM with detailed instructions and representative examples to demonstrate the extraction process. This approach eliminates the time-consuming and labor-intensive need for manual annotation and specific model fine-tuning, making it easily adaptable to various clinical questions \cite{llm-ner-medical,llm-information-medical}. We select Baichuan2-7B-Chat \cite{model-baichuan2-2023} to identify medical entities, which is an advanced open-source LLM and demonstrated excellent performance across multiple Chinese tasks. More details can be found in the code repository\footnote{https://github.com/Dragon-Wu/ICP-for-Clinical-Reasoning}. 
% 
% As a medical examination covers many medical subjects and clinical scenarios, its questions and the corresponding analysis involve numerous medical concepts, that are uncommon in general domain corpus and have different writing styles compared to other fields. Therefore, it is challenging to identify by the general NLP model exactly. 

\subsection{Construction of Medical Knowledge Graph}
In this subsection, we construct a medical knowledge graph, $G(E, V)$, to encapsulate the relationship among various medical entities, where $V$ and $E$ denote the sets of $m$ nodes $\{e_1,...,e_m\}$ and edges $v_{ij}$ of $G$, respectively. 
Each node in this graph represents a medical entity.
The directed edge $v_{ij}$ indicates the likelihood of discussing $e_j$ within an analysis given the presence of $e_i$ in the question and options. It is mathematically represented as:
\begin{equation}
v_{ij}=P(e_j\in R\ |\ e_i\in\{Q,O\})
\end{equation}
% therefore it is not the same as the $e_{ji}$, which shows the reverse opposing situation. 

An effective approach to compute the value of edge $v_{ij}$ is counting the instances of $e_i$ in $Q$ or $O$ and $e_j$ in the corresponding analysis $R$. Without any training, this process can be directly calculated based on a training dataset encompassing $n$ instances:
\begin{equation}
    \hat{v}_{ij}=\sum_{k=0}^{n}f(e_i,e_j,Q_k,O_k, R_k)
    \label{eqn:count}
\end{equation}
where $\{Q_k, O_k, R_k, A_k\}$ denotes a training sample $k$ (refer to section 3.1).  $f$ is defined as:
\begin{equation}
\begin{aligned}
    f(e_i,e_j,Q_k,&O_k,R_k) = \\
    &\left\{
        \begin{array}{ll}
        1, & if\ e_i\in\{Q_k,O_k\}\ and\ e_j\in R_k \\
        0, & otherwise \\
        \end{array}
    \right.
\end{aligned}
\end{equation}

Notably, due to the inherent directionality of the edges, $v_{ij}$ and $v_{ji}$ delineate the inverse path, leading to distinct values. If no relationship is established between node $e_i$ and node $e_j$, the $\hat{v}_{ij}$ is designated a value of -1.

Some common entities have been frequently mentioned in many questions. To enhance the specificity between associated entities, we implement a weighting scheme. The resultant weighted edge value, $v_{ij}$, can be formulated by:
\begin{equation}
    v_{ij} = \frac{\hat{v}_{ij}}{ \sum_{k=1}^{m}\hat{v}_{ik}} * \lg [\frac{m}{1+c_j}]
\end{equation}
% \\ m\ne i
Here, $c_j$ denotes the frequency with which entity $e_j$ appears in the analysis of the training dataset:
\begin{equation}
    c_j = \sum_{i=1}^{n} g(e_j, R_k)
\end{equation}
\begin{equation}
    g(e_j, R_k)=
    \left\{
        \begin{array}{ll}
        1, & if \ e_j \in R_k \\
        0, & otherwise \\
        \end{array}
    \right.
\end{equation}

To represent the entities and edges associated with $e_i$, we introduce the $V_{e_i}\in V$ and $E_{e_i}\in E$ to denote all edges sourced from $e_i$ and their targeted nodes. 

\subsection{Mining Potential Knowledge Seeds}
The essence of clinical context information can be approximately captured by the identified entities within $Q$ and $O$. 
We extract the entities $X$ from $\{Q, O\}$. 
For each extracted entity $x_i \in X$, the edges started from it $V_{x_i}$ and the targeted nodes $E_{x_i}$ are identified. 
Consequently, the aggregated sets of connected entities and edges are:
\begin{equation}
    E_{X}=\{ E_{e_1} \cup E_{e_2}, \cup ..., \cup E_{e_i}\}
\end{equation}
\begin{equation}
    V_{X}=\{ V_{e_1} \cup V_{e_2}, \cup ..., \cup V_{e_i}\}
\end{equation}
For each targeted entity $e \in E_{X}$ and its corresponding edge $v \in V_{X}$, the correlation of $e$ to an entity $x_i$ present in the question or options is computed. 
This relevance is determined by locating $e$ in $E_{x}$ and subsequently discerning the position of $v$ in the sorted $V_x$:
\begin{equation}
    q(e, v, E_x, V_x) = 
    \left\{
    \begin{array}{ll}
    rank(v, V_{x}), & if\ e \in E_{x} \\
    inf, & otherwise \\
    \end{array}
    \right.
\end{equation}
Subsequently, we accumulate the ranks of $e$ concerning all entities in $X$ to estimate the overall correlation of $e$ to all entities mentioned in $Q$ and $O$:
\begin{equation}
    q(e, X) = \sum_{i=1}^{} q(e, v, E_{x_i}, V_{x_i})
\end{equation}
Finally, we prioritize the entities $e \in E_X$ based on their value of $q(e, X)$ and pinpoint the top 10 entities as the \textit{knowledge seeds}, which are most likely to be discussed in inferences. 
% ranked ones

\subsection{Guiding Clinical Reasoning via Knowledge Seeds}
After detecting the knowledge seeds, we incorporate them into the prompt to steer the LLM toward producing a reasonable analysis. 
Specifically, we append these identified knowledge seeds following the questions and options. We then instruct the LLM with the detailed description, "\textit{Here is a clinical question, please refer to the knowledge seeds related to question-solving, and analyze this question step by step.}" This ensures that the LLM focuses and deliberates upon these entities, resulting in more concrete inference.

While we underscore the significance of these knowledge seeds, the LLM might encounter challenges in understanding or effectively using them. Moreover, certain entities may be solely related to the entities within $Q$ and $O$ but do not facilitate its solution. 
To efficiently guide the LLM in harnessing these knowledge seeds, we adopt a few-shot learning strategy. 
Representative samples in the format of \textit{\{question, option, knowledge seeds, analysis, and answer\}} are provided. 
Consequently, these exemplars guide the LLM to recognize helpful seeds and progressively factor them into the inferences, yielding a reasonable analysis and correct answer. 

\begin{table}[!tp]
\vspace{-0.36cm} 
\begin{center}{
    \resizebox{\linewidth}{!}{
    \begin{tabular}{ c  c  c  c  c}
        \toprule
        \multicolumn{1}{c}{\multirow{2}{*}{\textbf{Statistics}}} & \multicolumn{2}{c}{\textbf{CNMLE-Clinical}} & \multicolumn{2}{c}{\textbf{CMExam}} \bigstrut\\
        \cline{2-5}  & \multicolumn{1}{c}{\textbf{Train}} & \multicolumn{1}{c}{\textbf{Test}} & \multicolumn{1}{c}{\textbf{Train}} & \multicolumn{1}{c}{\textbf{Test}} \bigstrut\\
        % \cline{1-5}
        % Number of questions  &14655 	&600 	&56965	&600 \\
        \cline{1-5}
        \multirow{2}{*}{Number of questions} & \multirow{2}{*}{14655} & \multirow{2}{*}{600} & \multirow{2}{*}{56965}	& \multirow{2}{*}{600} \\
        \multirow{2}{*}{} & & & & \\
        \cline{1-5}
        \multirow{1}{*}{Average length of} & \multirow{2}{*}{101.23} & \multirow{2}{*}{102.71} & \multirow{2}{*}{85.56}	& \multirow{2}{*}{86.70} \\
        \multirow{1}{*}{question and options} & & & & \\
        \cline{1-5}
        \multirow{1}{*}{Average length of }	& \multirow{2}{*}{89.26} & \multirow{2}{*}{96.54} & \multirow{2}{*}{220.02}	& \multirow{2}{*}{208.82} \\
        \multirow{1}{*}{analysis} & & & & \\
        \cline{1-5}
        \multirow{1}{*}{Number of entity} & \multirow{2}{*}{14.57} & \multirow{2}{*}{14.79} & \multirow{2}{*}{13.57}	& \multirow{2}{*}{13.11} \\
        \multirow{1}{*}{within question and options} & & & & \\
        \cline{1-5}
        \multirow{1}{*}{Number of entity} & \multirow{2}{*}{11.39} & \multirow{2}{*}{12.56} & \multirow{2}{*}{29.09}	& \multirow{2}{*}{25.33} \\
        \multirow{1}{*}{within analysis} & & & & \\
        \bottomrule
    \end{tabular}
    }
    }
\end{center}
\vspace{-0.25cm} 
\caption{Dataset Statistics.}
\vspace{-0.25cm} 
\label{Table:statistics}
\end{table}

\section{Experiments and Results}
\subsection{Dataset}
The Chinese National Medical Licensing Examination (CNMLE)\footnote{https://www1.nmec.org.cn/Pages/ArticleInfo-13-10706.html} serves as the official certification examination for medical practitioners in China, akin to the United States Medical Licensing Examination (USMLE). 
% Annually, more than half a million medical practitioners participate in the CNMLE. 
It is a prerequisite for the candidates to have undergone five years of medical education, in addition to at least one year of assessed clinical practice.
The objective of the CNMLE is 
% not only to assess the proficiency of medical knowledge of the candidates but also 
to evaluate the practical abilities in a real-world clinical setting. Therefore, we built two clinical reasoning data sets from CNMLE.
% The CNMLE comprises various types of questions, which can be entirely transformed into objective multiple-choice questions, each offering five alternative options. 
% % The CNMLE is divided into two main sections: practical skills examination and comprehensive medical examination. 
% % Our research primarily focuses on the latter. 
% An examination includes 600 questions in total. To pass the CNMLE, candidates must correctly answer at least 60\%(360) of these questions, thus achieving a minimum score of 60.
% To comprehensively evaluate the performance of different methods, we adopt two datasets related to CNMLE to conduct experiments. 
The dataset statistic is shown in Table \ref{Table:statistics}.

\subsubsection{CNMLE-Clinical}
There are different examinations for different clinical disciplines, the first dataset \textit{CNMLE-Clinical} focuses on CNMLE for clinical medicine. 
The CNMLE-Clinical evaluates four parts of medicine: preventive medicine, preclinical medicine, clinical medicine, and medical humanities, which cover over twenty distinct medical subjects. 
We gathered questions from past examinations and various reference books, accumulating a total of 15,255 questions. Out of these, we randomly selected 600 questions, consistent with the number in the official examination, to serve as our testing set, while the remainder were used as the training set. 
Each instance in CNMLE-Clinical consists of a question, five candidate options, the correct answer, and a detailed explanation for the answer.

% \begin{table*}[htbp]\centering
%   \caption{Performance on Dataset CMExam.}
%   \scriptsize
%   % \resizebox{0.86\linewidth}{!}{
%   \begin{tabular}{lcccccccc}
%     \toprule
%     \textbf{Method} &\makecell[c]{\textbf{Acc}(\%)} &\makecell[c]{\textbf{BLEU-1}} &\makecell[c]{\textbf{BLEU-2}} &\makecell[c]{\textbf{BLEU-3}} &\makecell[c]{\textbf{BLEU-4}} &\makecell[c]{\textbf{ROUGE-1}} &\makecell[c]{\textbf{ROUGE-2}} &\makecell[c]{\textbf{ROUGE-L}} \\
%     \midrule
%     \textbf{Zero-shot} & & & & & & & & \\
%     Standard QA & \textbf{45.83}\% & / & / & / & / & / & / & / \\
%     Chain-of-Thought & 42.67\% & 26.36  & 18.62  & 14.49  & 11.36  & 36.54  & 11.94  & 18.53 \\
%     ICP (Our method) & 44.83\% & \textbf{29.02} & \textbf{20.83}  & \textbf{16.35}  & \textbf{12.85}  & \textbf{35.76}  & \textbf{12.40}  & \textbf{20.32} \\
%     \midrule
%     \textbf{Few-shot} & & & & & & & & \\
%     Standard QA & 47.33\% & / & / & / & / & / & / & / \\
%     Chain-of-Thought & 50.00\% & 31.04  & 23.74  & 19.29  & 15.79  & 40.87  & 18.97  & 26.85 \\
%     ICP (Our method) & \textbf{51.33}\% & \textbf{32.83}  & \textbf{25.31}  & \textbf{20.69}  & \textbf{17.00}  & \textbf{42.08}  & \textbf{20.08}  & \textbf{27.78} \\
%     \bottomrule
%     \end{tabular}
%     \label{table: Performance cmexam}
%     % }
% \end{table*}

\begin{table*}[htbp]\centering
  \vspace{-0.16cm} 
  \tiny
  % \resizebox{0.86\linewidth}{!}{
  \begin{tabular}{lcccccccc}
    \toprule
    \textbf{Method} & \textbf{Acc(\%)} & \textbf{BLEU-1} & \textbf{BLEU-2} & \textbf{BLEU-3} & \textbf{BLEU-4} & \textbf{ROUGE-1} & \textbf{ROUGE-2} & \textbf{ROUGE-L} \\
    \midrule
    \textbf{PLM-Finetuned} & \textbf{} & \textbf{} & \textbf{} & \textbf{} & \textbf{} & \textbf{} & \textbf{} & \textbf{} \\
    Bert & 31.80\% & - & - & - & - & - & - & - \\
    RoBERTa & 37.10\% & - & - & - & - & - & - & - \\
    Bart-base & - & 23.00 & - & - & 10.35 & 44.33 & 24.29 & 20.80 \\
    Bart-large & - & 26.37 & - & - & 11.65 & 44.92 & 24.34 & 21.75 \\
    PromptCLUE (T5) & - & 18.75 & - & - & 6.65 & 40.88 & 21.90 & 18.31 \\
    \midrule
    \textbf{LLM-Finetuned} &  &  &  &  &  &  &  &  \\
    LLaMA & 18.30\% & 29.25 & - & - & 16.46 & 45.88 & 26.57 & 23.31 \\
    ChatGLM & 45.30\% & 31.10 & - & - & 18.94 & 43.94 & 31.48 & 29.39 \\
    Huatuo & 28.60\% & 29.04 & - & - & 16.72 & 43.85 & 25.36 & 21.72 \\
    MedAlpaca & 30.05\% & 16.35 & - & - & 9.78 & 44.31 & 27.05 & 24.55 \\
    \midrule
    \textbf{LLM-Zero-shot} & \textbf{} & \textbf{} & \textbf{} & \textbf{} & \textbf{} & \textbf{} & \textbf{} & \textbf{} \\
    Standard QA & 45.83\% & - & - & - & - & - & - & - \\
    Chain-of-Thought & 42.67\% & 34.02 & 24.27 & 18.03 & 13.57 & 46.20 & 20.39 & 21.97 \\
    ICP (Our method) & 44.83\% & 42.73 & 30.43 & 22.57 & 16.89 & 45.24 & 20.84 & 23.40 \\
    \midrule
    \textbf{LLM-Few-shot} &  &  &  &  &  &  &  &  \\
    Standard QA & 47.33\% & - & - & - & - & - & - & - \\
    Chain-of-Thought & 50.00\% & 58.38 & 42.30 & 32.09 & 24.69 & 48.86 & 26.39 & 28.88 \\
    ICP (Our method) & \textbf{51.33}\% & \textbf{59.84} & \textbf{43.68} & \textbf{33.38} & \textbf{25.86} & \textbf{49.89} & \textbf{27.43} & \textbf{29.70} \\
    \bottomrule
    \end{tabular}
    \caption{Performance on Dataset CMExam.}
    \label{table: Performance cmexam}
    % }
    \vspace{-0.36cm} 
    
\end{table*}

\subsubsection{CMExam}
The second dataset, \textit{CMExam} \cite{dataset-cmexam-2023}, is a more comprehensive collection encompassing the six types of medical licensing examination: clinical medicine, traditional Chinese medicine (TCM), integrated TCM and Western medicine, dentistry, public health, pharmacy, and traditional Chinese pharmacy. 
The CMExam incorporates 68,119 medical questions sourced from past examinations and medical books. 
% These questions were primarily gathered from the Internet, with some lacking an analysis or existing in wrong format. 
To ensure consistency with the genuine examination, we included 57,565 questions that have five options, a singular correct answer, and an analysis exceeding 30 words. 
Lastly, we randomly selected 600 questions to serve as the testing set, while the remaining questions in CMExam were designated as the training set.
As shown in Table \ref{Table:statistics}, the analyses of CMExam usually have more details than CNMLE-Clinical, which contains 25-29 entities in analyses, significantly higher than CNMLE-Clinical.

\subsection{Settings}
The GPT 3.5-Turbo is selected as the primary model to evaluate, which includes 175B parameters and drives the online ChatGPT. All tests were conducted by calling OpenAI’s official API. Unless specified, all experiments used the same parameters and were tested with the same version of the model (gpt-3.5-turbo-0613). To make the responses more deterministic and repeatable, we set the inference temperature to 0 to conduct greedy decoding, which always chooses the token with maximum probability instead of sampling.
We keep the maximum context length (including prompt and response) of GPT 3.5-Turbo (4097 tokens) to avoid the potential performance penalty due to response length. The rest parameters are all set to default.

\subsection{Evaluation Metrics}
We evaluate model performance using both accuracy and NLG metrics. The accuracy generally represents the model's ability to conduct clinical reasoning. 
To assess the quality of the generated analysis, we employ BLUE \cite{metric-bleu-2002} and ROUGE \cite{metric-ROUGE-2004}, which are commonly used in medical inference \cite{dataset-cmexam-2023,google-multimodality-2023} or medical report generation \cite{cvpr-chexpert-2021}.

\subsection{Baselines}
To fully reveal the performance of LLMs with ICP, we evaluate several competitive baselines as follows. 
% \begin{itemize}
% \item \textbf{Standard QA} This baseline employs the standard question-and-answer(\textbf{QA}) prompting that requires the LLM to respond directly to the predicted answer without any explanation. The prompt is "\textit{Here is a multi-choice question about medical knowledge, please output the correct answer according to the question.}"
% \item \textbf{Chain-of-Thought (CoT)} promoting was introduced by \cite{cot-first-wei2022} and \cite{cot-step-kojima2022}, it guides the LLM to solve questions step by step and prompts it to generate a detailed reasoning process before reaching the final conclusions. The prompt is "\textit{Here is a multi-choice question about medical knowledge, please analyze it in a step-by-step fashion and deduce the correct answer.}" 
% \end{itemize}
% Furthermore, to enhance the inference capabilities of LLMs, we leverage in-context learning through the few-shot approach. We examine both \textbf{Zero-shot} and \textbf{Few-shot} strategies. Due to the context length limitations of LLMs, we use six examples in the few-shot experiments. To encompass various question types, these six examples cover clinical case analysis, comprehension of medical knowledge, and medical computation tasks.

\begin{itemize}
    \item \textbf{PLM-Finetuned:} These baselines involve pre-trained language models (PLMs) extensively fine-tuned on the training set, including \textbf{BERT} \cite{plm-bert}, \textbf{RoBERTa} \cite{plm-roberta}, \textbf{Bart-base} \cite{plm-bart}, \textbf{Bart-large}, and \textbf{PromptCLUE} \cite{plm-promptclue} (based on T5 model \cite{plm-t5}). 
    Encoder-only models like BERT and RoBERTa are designed to output results directly, whereas encoder-decoder models such as Bart and PromptCLUE can generate detailed explanations.
    Experimental setup and results referred to \citeauthor{dataset-cmexam-2023}(\citeyear{dataset-cmexam-2023}). 

    \item \textbf{LLM-Finetuned:} These baselines include LLMs fully fine-tuned on the training set, including both general LLMs (\textbf{LLaMA} \cite{llm-llama} and \textbf{ChatGLM} \cite{llm-glm}) and medical LLMs (\textbf{Huatuo} \cite{llm-huatuo} and \textbf{MedAlpaca} \cite{llm-medalpaca}). They support the generation of inferences and final answers. Moreover, the medical LLMs have also undergone additional training on medical corpora and medical question datasets \cite{llm-huatuo,llm-medalpaca}. 
    % Experimental setup and results are detailed in \citeauthor{dataset-cmexam-2023}(\citeyear{dataset-cmexam-2023}).
    
    \item \textbf{LLM-Zero-shot:} The \textbf{Standard QA} employs the standard question-and-answer prompting that instructs the LLM (GPT3.5) to respond directly to the predicted answer without any explanation, as "\textit{Here is a multi-choice question about medical knowledge, please output the correct answer according to the question.}" 
    The \textbf{Chain-of-Thought (CoT)} promoting was introduced by \cite{cot-first-wei2022} and \cite{cot-step-kojima2022}, which guides the LLM to solve questions step by step and prompts it to generate a detailed analysis, as "\textit{Here is a multi-choice question about medical knowledge, please analyze it in a step-by-step fashion and deduce the correct answer.}" 

    \item \textbf{LLM-Few-shot:} We further enhance the reasoning capabilities of LLM through \textbf{Few-shot} approach, leveraging the in-context learning ability of LLMs. Given the context length constraints of LLMs, we randomly select six examples across a variety of clinical question types, including clinical case analysis, understanding of clinical knowledge, and medical computation tasks.
    
\end{itemize}

\begin{table*}[htbp]\centering
  % \vspace{-0.16cm} 
  \tiny
  % \resizebox{0.80\linewidth}{!}{
  \begin{tabular}{lcccccccc}
    \toprule
    \textbf{Method} &\makecell[c]{\textbf{Acc}(\%)} &\makecell[c]{\textbf{BLEU-1}} &\makecell[c]{\textbf{BLEU-2}} &\makecell[c]{\textbf{BLEU-3}} &\makecell[c]{\textbf{BLEU-4}} &\makecell[c]{\textbf{ROUGE-1}} &\makecell[c]{\textbf{ROUGE-2}} &\makecell[c]{\textbf{ROUGE-L}} \\
    \midrule
    \textbf{LLM-Zero-shot} & & & & & & & & \\
    Standard QA  & 51.00\% & / & / & / & / & / & / & / \\
    Chain-of-Thought & 48.00\% & 12.47 & 8.75 & 6.32 & 4.57 & 37.31 & 15.60 & 16.76 \\
    ICP (Our method) & \textbf{53.33\%} & \textbf{19.38} & \textbf{13.72} & \textbf{10.01} & \textbf{7.33} & \textbf{38.43} & \textbf{16.84} & \textbf{19.54} \\
    \midrule
    \textbf{LLM-Few-shot} & & & & & & & & \\
    Standard QA & 51.83\% & / & / & / & / & / & / & / \\
    Chain-of-Thought & 54.83\% & 35.72 & 25.29 & 18.46 & 13.49 & 41.68 & 18.91 & 23.88 \\
    ICP (Our method) & \textbf{58.83\%} & \textbf{35.78} & \textbf{25.47} & \textbf{18.69} & \textbf{13.74} & \textbf{42.21} & \textbf{19.47} & \textbf{24.28}\\
    \bottomrule
    \end{tabular}
    \caption{Performance on Dataset CNMLE-Clinical.}
    \label{table: Performance cnmle}
    % }
    % \vspace{-0.1cm} 
\end{table*}

\begin{table*}[htbp]
    \centering
    \vspace{-0.16cm} 
    \tiny
  % \resizebox{0.86\linewidth}{!}{
  \begin{tabular}{lcccccccc}
    \toprule
    \textbf{Method} &\makecell[c]{\textbf{Acc}(\%)} &\makecell[c]{\textbf{BLEU-1}} &\makecell[c]{\textbf{BLEU-2}} &\makecell[c]{\textbf{BLEU-3}} &\makecell[c]{\textbf{BLEU-4}} &\makecell[c]{\textbf{ROUGE-1}} &\makecell[c]{\textbf{ROUGE-2}} &\makecell[c]{\textbf{ROUGE-L}} \\
    \midrule
    \textbf{Baichuan2-Chat (7B)} & & & & & & & & \\
    CoT & 42.83\% & 38.54 & 27.08 & 19.81 & 14.63 & 43.99 & 19.40 & 22.21 \\
    ICP & 44.67\% & 52.75 & 36.92 & 26.94 & 19.81 & 44.47 & 20.95 & 24.88 \\
    CoT with Few-shot & 49.83\% & 53.12 & 37.26 & 27.44 & 20.46 & 41.89 & 21.63 & 24.60 \\
    ICP with Few-shot & \textbf{50.00\%} & \textbf{56.10} & \textbf{39.95} & \textbf{29.79} & \textbf{22.44} & \textbf{45.91} & \textbf{24.53} & \textbf{26.50} \\
    \midrule
    \textbf{GPT-4 ($\sim$ 1.7T)} & & & & & & & & \\
    CoT & 72.00\% & 38.73 & 27.43 & 20.16 & 14.94 & 45.91 & 19.34 & 21.88 \\
    ICP & 72.00\% & 43.21 & 31.13 & 23.24 & 17.50 & 47.25 & 21.99 & 24.56 \\
    CoT with Few-shot & 76.00\% & 56.59 & 41.34 & 31.82 & 24.98 & 50.41 & 28.29 & 29.79 \\
    ICP with Few-shot & \textbf{78.00\%} & \textbf{60.15} & \textbf{44.34} & \textbf{34.44} & \textbf{27.25} & \textbf{51.87} & \textbf{29.99} & \textbf{31.66} \\
    \bottomrule
    \end{tabular}
    \caption{Performance of Different Base LLMs on CMExam.}
    \label{table: bc2 and gpt4}
    % }
    \vspace{-0.36cm} 
\end{table*}

\subsection{Result}
As illustrated in Table \ref{table: Performance cmexam} and Table \ref{table: Performance cnmle}, the proposed ICP outperforms baselines in both CMExam and CNMLE-Clinical, which significantly improves both the accuracy and quality of generated analysis. This verified the identified knowledge seeds encompass crucial information and thus efficiently guide the LLM to generate a convincing inference. Concurrently, we observe a strong coherence between accuracy and NLG metrics, which suggests that an improvement in the reasoning process aids in addressing clinical questions. 

\paragraph{Effect of Generating Analysis.} Compared to the QA-based method, the CoT-based method, although producing detailed explanations, often compromises the final accuracy. This is different from the findings in other domains where generating a detailed reasoning process prior to outputting an answer can substantially improve model performance, as seen in mathematical reasoning \cite{cot-first-wei2022,cot-step-kojima2022}. This divergence in the medical domain might arise because of its intrinsic demand for precise medical knowledge. Without the guidance of reliable medical knowledge, CoT might inadvertently generate false knowledge or hallucinations, consequently impairing performance. Similarly, the previous evaluation of LLMs in medicine \cite{exam-chatgpt-2023} also verified such a phenomenon. 

\paragraph{Effect of Knowledge Seed.} In contrast to CoT, our introduced ICP framework also improves the quality of generated explanations. Under the zero-shot setting, ICP achieves an increment of 3.32 in BLEU-4 and 1.43 in ROUGE-L for CMExam. Similarly, for CNMLE-Clinical, the uplift is 2.73 in BLEU-4 and 2.78 in ROUGE-L, resulting in accuracy improvements of 2.16 and 5.33, respectively. 
In CMExam with zero-shot, the accuracy of ICP is slightly lower than QA. 
This can be attributed to the intricacies of questions in CMExam, which involve more disciplines and usually contain more entities in longer analysis than CNMLE-Clinical (see Table \ref{Table:statistics}). 
The incorporation of Few-shot strategy enables LLM to fully harness its in-context learning ability, which efficiently learns the patterns for problem-solving from examples, thus consistently improving model performance. The ICP with few-shot adeptly utilizes the predicted knowledge seeds for inference and achieves the best performance.

% ACC of ICP vs cot
% cmexam: zero-2.16, few-1.33 
%  cnmle: zero-5.33, few-4.00 
% NLG
% zero-shot
%         bleu-4  ROUGE-L
% cmexam: 3.32, 1.43
%  cnmle: 2.73, 2.78

% few-shot
%         bleu-4  ROUGE-L
% cmexam: 1.17, 0.82
%  cnmle: 0.25, 0.4

\section{Analysis and Case Study}
\subsection{Impact of the LLM Size}
To investigate the efficacy and generalizability of our framework, we conducted a comprehensive performance analysis on LLMs of varying sizes.

\paragraph{Baichuan2-chat (7B).} The Baichuan2-chat-7B (BC2) was chosen to represent small LLMs, with its 7 billion parameters making it suitable for deployment on common PCs. 
It is an open-source bilingual LLM (Chinese and English) and has exhibited fine performance across various tasks. 
We assessed the reasoning abilities of BC2 under both zero-shot and few-shot scenarios for CoT and ICP using CMExam. 
Given BC2's extensive training on Chinese data, it achieved excellent performance on CMExam, only slightly behind GPT-3.5 (7B vs 175B). 
In a few-shot setting with ICP, its performance was nearly on par (50 vs 51.33), with a ROUGE-L score closely matching GPT-3.5 (26.50 vs 29.70). 
Such consistent and substantial improvements in a 7B model further validate the superiority of our framework, highlighting the potential of applying it to more resource-limited regions. 
     
\paragraph{GPT-4 ($\sim$1.7T).} 
Subsequently, we examined GPT-4, which is unofficially estimated to possess around 1.7 trillion parameters, making it the most powerful model available in terms of both size and performance. We accessed the GPT-4 (version: 0613) via its official API with the same parameters as GPT-3.5. 
Due to the API access rate limitations, we randomly selected 50 samples for testing. 
The results underscored GPT-4's exceptional capabilities, as it achieved an accuracy of 72\% even in zero-shot scenarios. Moreover, ICP consistently enhanced the model's accuracy and inference capabilities. This improvement also proves that our method can further enhance the model's ability for larger LLMs.

\subsection{Performance on Different Medical Disciplines}
To explore the usability of the LLMs integrated with ICP in medical scenarios, we further investigate its performance across different medical disciplines, areas of competency, and clinical departments. The subsequent analyses are based on the result of ICP under a few-shot setting with GPT3.5 in CMExam, as shown in Tables \ref{table: different examination}, \ref{table: areas of competency}, and \ref{table: clinical department}.

As presented in Table \ref{table: different examination}, 
% the model performance varies significantly across different examinations. 
For clinical medicine, LLM with ICP achieved a high score of 60.77, qualifying for this licensing examination. This implies that its proficiency in medical knowledge is comparable to that of one certified physician. Furthermore, it achieved the highest ROUGE-L at 32.78 and BLEU-4 at 29.09. However, GPT-3.5 encounters challenges in public health and preventive medicine, yielding the lowest accuracy of 30.77\%. In contrast, BC2 achieved an accuracy of 50\%. Given that GPT-3.5 is predominantly trained on English corpus, this disparity is primarily attributed to that it involves many Chinese healthcare policies and regulations, such as "\textit{What is the stipulated standard for total coliform in China's drinking water sanitation guidelines?}".

\begin{table}[tp]
    \centering
    \resizebox{\linewidth}{!}{
    \begin{tabular}{ccccc}
        \toprule
        \textbf{Clinical Discipline} & \textbf{Count} & \textbf{Accuracy} & \textbf{ROUGE-L} & \textbf{BLEU-4} \\
        \midrule
        Clinical medicine & 260 & \textbf{60.77\%} & \textbf{32.78} & \textbf{29.09} \\
        Dentistry & 95 & 43.16\% & 29.04 & 23.85 \\
        Pharmacy & 74 & 54.05\% & 29.45 & 26.94 \\
        Traditional Chinese pharmacy & 65 & 40.00\% & 23.44 & 21.12 \\
        Traditional Chinese medicine(TCM) & 31 & 41.94\% & 26.06 & 20.74 \\
        Integrated TCM and Western medicine & 49 & 44.90\% & 26.91 & 22.80 \\
        Public health and preventive medicine & 26 & 30.77\% & 27.32 & 21.51 \\
        \bottomrule
    \end{tabular}
    }
    \vspace{-0.15cm}
    \caption{The Performance on Different Clinical Disciplines.}
    \label{table: different examination}
\end{table}

\begin{table}[tp]
    \vspace{-0.15cm}
    \centering
    \resizebox{\linewidth}{!}{
    \begin{tabular}{ccccc}
    \toprule
    \textbf{Area of Competency} & \textbf{Count} & \textbf{Accuracy} & \textbf{ROUGE-L} & \textbf{BLEU-4} \\
    \midrule
    Disease diagnosis & 221 & 53.85\% & 30.23 & \textbf{26.86} \\
    Medical knowledge & 197 & \textbf{54.31\%} & 30.17 & 25.58 \\
    % \midrule
    Disease treatment & 135 & 43.70\% & 27.86 & 24.91 \\
    % \midrule
    Preventive medicine & 40 & 52.50\% & \textbf{30.57} & 24.55 \\
    \bottomrule
    \end{tabular}
    }
    \vspace{-0.15cm}
    \caption{Performance on Different Areas of Physician Competency.}
    \label{table: areas of competency}
\end{table}

To holistically evaluate a physician's capabilities, the examination integrates a variety of clinical questions, including an assessment of medical knowledge, disease diagnosis, disease treatment, and preventive medicine. As illustrated in Table \ref{table: areas of competency}, the LLM achieves an accuracy exceeding 52\% in three areas. The disease treatment necessitates considerations of the main symptoms, disease history, current condition, and other factors, making it a more complex clinical reasoning. Therefore, more adaptations and improvements are required to cope with these questions. 

\begin{table}[!tp]
    \vspace{-0.15cm}
    \centering
    \resizebox{\linewidth}{!}{
    \begin{tabular}{ccccc}
        \toprule
        \textbf{Clinical Department} & \textbf{Count} & \textbf{Accuracy} & \textbf{ROUGE-L} & \textbf{BLEU-4} \\ 
        \midrule
        Internal medicine & 118 & 56.78\% & 30.98 & 28.21 \\
        TCM & 90 & 38.89\% & 23.68 & 19.47 \\
        General & 67 & 50.75\% & 30.46 & 25.95 \\
        Dentistry & 50 & 44.00\% & 30.98 & 26.81 \\
        Surgery & 43 & 48.84\% & 32.92 & 28.95 \\
        Clinical laboratory & 35 & \textbf{71.43}\% & 28.91 & 25.36 \\
        Infectious diseases & 31 & 51.61\% & 28.55 & 24.51 \\
        Obstetrics and gynecology & 26 & 57.67\% & 29.77 & 27.15 \\
        Pediatrics & 19 & 63.16\% & 31.17 & 26.67 \\
        Radiology & 16 & 43.75\% & \textbf{35.44} & \textbf{30.05} \\
        \bottomrule
    \end{tabular}
    }
    \vspace{-0.15cm}
    \caption{ Performance on Different Clinical Departments.}
    \label{table: clinical department}
    \vspace{-0.16cm}
\end{table}

Diving deeper into various clinical departments based on Table \ref{table: clinical department}, it reveals that the model performance varies considerably across diseases, ranging from a low of 38.89\% in TCM to a high of 71.43\% in the clinical laboratory. This variation can be attributed to the inherent complexity differences among diseases; for instance, questions about the laboratory department primarily focus on the interpretation of biomarkers, offering a more concentrated context. 
% Additionally, the distribution of department-specific corpus during the LLM development might influence the encoded knowledge in various diseases. 

\subsection{Error Analysis}
To further investigate the role of the ICP framework in the reasoning process, we conducted an error analysis based on the results of ICP with Few-shot (GPT-3.5) in CMExam, as recorded in Table \ref{table: comparison of analysis}. 
We analyzed the metrics associated with the predicted knowledge seeds. Each ground-truth explanation contains nearly ten medical entities. When compared with our predicted set of ten knowledge seeds, it became evident that questions with correct answers often had better knowledge seeds, as indicated by higher recall, precision, and F1 scores.
Furthermore, regarding the length of responses and NLG metrics, we observed that the correct response typically has higher ROUGE and BLEU scores despite having a shorter average length compared to the incorrect one. 
% This indicates that with more accurate knowledge seeds, the reasoning process can focus on critical aspects, leading to effective discussions and correct answers. 
Overall, this suggests that our designed ICP framework and mined knowledge seeds can guide LLMs to conduct focused discussions on potentially critical aspects for question-solving, thereby effectively enhancing the quality of generated analyses and improving their reasoning ability in the medical domain.

% This suggests that our designed knowledge seeds can effectively elevate the quality of generated explanations, subsequently improving the problem-solving ability of LLM in medicine. 

\begin{table}[!tp]
    \centering
    \resizebox{\linewidth}{!}{
    \begin{tabular}{cccccccc}
    \toprule
    \multirow{2}{*}{} & \multirow{2}{*}{\textbf{\begin{tabular}[c]{@{}c@{}}Count \\ of KS\end{tabular}}} & \multirow{2}{*}{\textbf{\begin{tabular}[c]{@{}c@{}}Precision \\ of KS\end{tabular}}} & \multirow{2}{*}{\textbf{\begin{tabular}[c]{@{}c@{}}Recall \\ of KS\end{tabular}}} & \multirow{2}{*}{\textbf{\begin{tabular}[c]{@{}c@{}}F1-score \\ of KS\end{tabular}}} & \multirow{2}{*}{\textbf{Length}} & \multirow{2}{*}{\textbf{ROUGE-L}} & \multirow{2}{*}{\textbf{BLEU-4}} \\
     &  &  &  &  &  &  &  \\
    \midrule
    Correct & 9.88 & \textbf{0.312} & \textbf{0.285} & \textbf{0.266} & 137.48 & \textbf{31.09} & \textbf{27.21} \\
    \midrule
    Incorrect & 10.75 & 0.283 & 0.272 & 0.249 & 141.08 & 28.24 & 24.42 \\
    \bottomrule
    % \normalsize (The bold means the recommended metric)
    \end{tabular}
    }
    \vspace{-0.15cm}
    \caption{Comparison of Generated Analysis by Answering Questions Correctly and Incorrectly. KS means knowledge seeds}
    \label{table: comparison of analysis}
    \vspace{-0.36cm} 
\end{table}

\begin{figure}[!tp]
    \centering
        \includegraphics[width=0.43\textwidth]{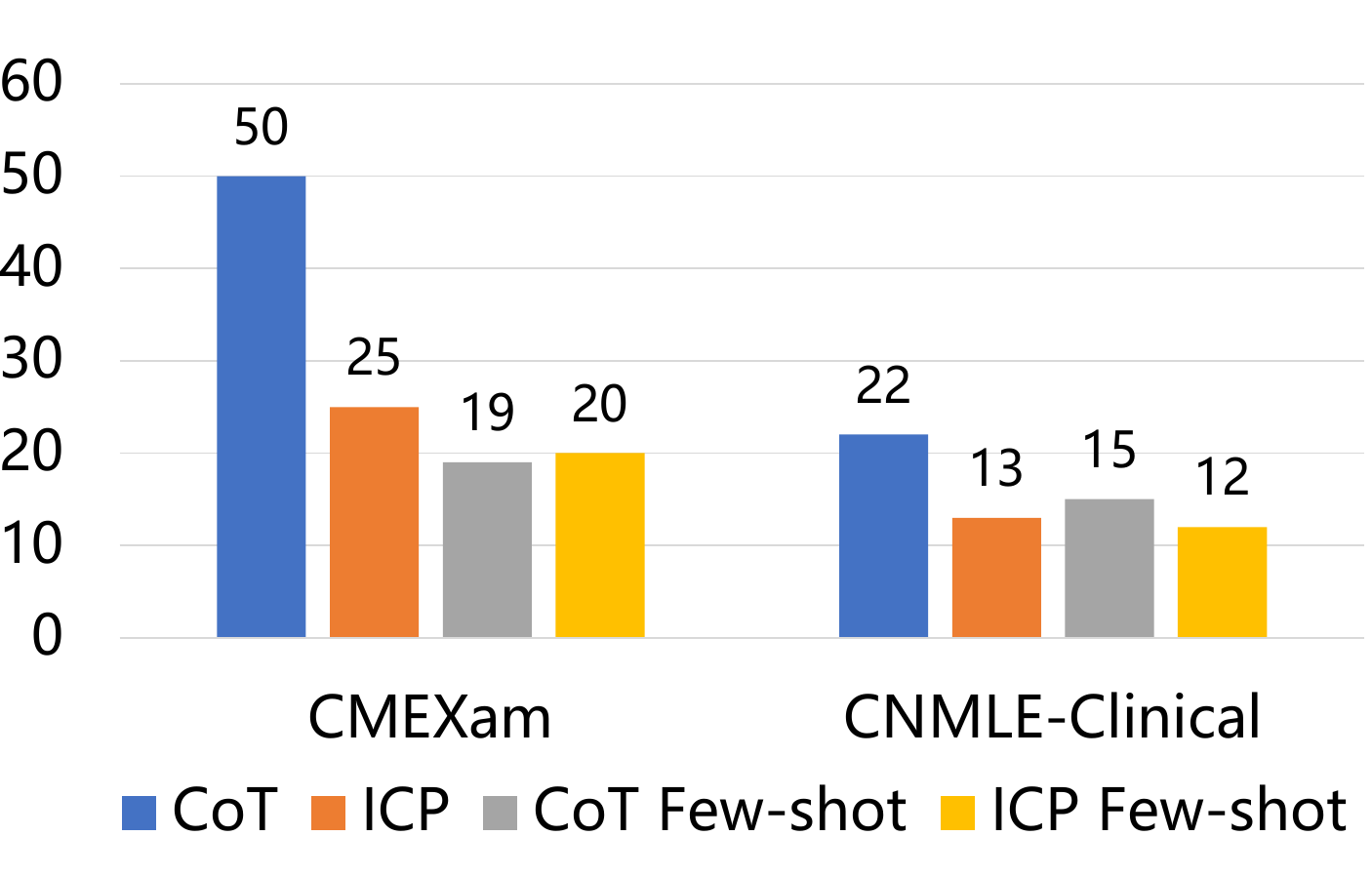}
    \vspace{-0.15cm}
    \caption{Questions Failed to Conclude An Answer.}
    \label{fig:fail}
    \vspace{-0.36cm} 
\end{figure}

After reviewing the reasoning process, unlike the standard QA that directly provides an answer option, several analyses of CoT and ICP fail to conclude with an exact answer option, as highlighted in Figure \ref{fig:fail}. It suggests that there might be multiple correct answers or no correct option. While there are insightful discussions in the reasoning process, and some approximately reached the answer, the final answer is still uncertain due to the deficiency of medical knowledge or incomplete reasoning. 
This underscores the rigor and complexity of clinical reasoning, necessitating further exploration into integrating more knowledge or enhancing reasoning abilities.

% \begin{table}[htbp]
%     \centering
%     \caption{Questions Failed to Conclude An Answer.}
%     \resizebox{\linewidth}{!}{
%     \begin{tabular}{ccccc}
%     \toprule
%     \multirow{2}{*}{} & \multirow{2}{*}{\textbf{CoT}} & \multirow{2}{*}{\textbf{ICP}} & \multirow{2}{*}{\textbf{\begin{tabular}[c]{@{}c@{}}CoT \\ Few-shot\end{tabular}}} & \multirow{2}{*}{\textbf{\begin{tabular}[c]{@{}c@{}}ICP \\ Few-shot\end{tabular}}} \\
%      &  &  &  &  \\
%     \midrule
%     \textbf{CMEXam} & 50 & 25 & 19 & 20 \\
%     \midrule
%     \textbf{CNMLE-Clinical} & 22 & 13 & 15 & 12 \\
%     \bottomrule
%     \end{tabular}
%     }
%     \label{table: no answer}
% \end{table}

% \subsection{Case Study}
% some examples

\section{Conclusion}
% We proposed a simple yet effective in-context padding framework that mines and inserts external knowledge seeds into the prompt.
In this study, we proposed a simple yet effective In-Context Padding framework that 1) identifies potential knowledge seeds using the medical knowledge graph; and 2) enhances LLM on clinical reasoning. 
Experiments have shown that our framework can significantly enhance the inference capabilities of LLMs. During the LLM reasoning process, the mined knowledge seeds effectively bridge the knowledge gaps between LLMs and the medical domain. Extensive ablation studies and error analysis have proven the robustness and generalizability of our framework. 
Our efforts aim to ensure that LLMs are effective and equitable in specialized domains, especially in healthcare, thereby promoting global health equity through artificial intelligence solutions.

% \section*{Limitations}
% Several potential limitations should be considered for this study. Firstly, the established KG relies on the pairs of question-analysis in the training set. This might constrain the scale of the KG and then limit its extension to more different datasets. In future work, we will explore how to efficiently construct a comprehensive KG by incorporating more external knowledge sources, such as medical literature and Wikipedia. Secondly, due to time and resource constraints, we only employed prompts written in Chinese to conduct medical inference and did not test for prompts in English. However, the Chinese prompt is also consistent with actual medical examinations and practical applications. 

\section*{Ethical Statement}
While the explored clinical questions involve specific analysis of clinical cases, all cases have been anonymized and de-identified to ensure that no personal information is disclosed
The primary objective of this research is to investigate the reasoning abilities of LLMs in the clinical domain. To that end, we opted for a comprehensive evaluation using clinical questions. 
However, the results and discussions of this research are purely for academic research and analysis and do not serve as actual medical suggestions. Therefore, there is no negative impact on human health. 

% Timely and accurate clinical reasoning is crucial for effective healthcare delivery, imposing high demands on the knowledge and experience of physicians. However, in developing countries, the gap between the high medical demands and limited physician availability underscores the urgent need for automated clinical reasoning solutions to mitigate global health inequities. Although recent advances in large language models (LLMs) have shown potential in this area, they still encounter challenges like hallucinations. To address this, we propose a novel framework, In-Context Padding (ICP), to enhance the clinical reasoning capabilities of LLMs. Experimental results on two datasets of clinical questions demonstrated significant improvement in LLMs' ability to conduct complex clinical reasoning. Overall, this study contributes a practical approach to alleviating healthcare barriers, especially for under-resourced regions, promoting Good Health and Well-being for all humanity. 

% enabling more individuals to benefit from the rapid advancements in AI

% \appendix

\section*{Contribution Statement}
Jiageng Wu and Xian Wu contributed equally to this paper. Correspondence to Jie Yang. 

% \clearpage
\bibliographystyle{named}
\bibliography{ijcai24}

\end{document}

\maketitle
\section*{A. Prompt Setting}

\section*{B. Case Study}

\subsection*{B.1 Related Work}

% \appendix

% \clearpage
% \bibliographystyle{named}
% \bibliography{ijcai24}